\documentclass[conference]{IEEEtran}
\usepackage{hyperref}
\usepackage{url}

\usepackage{nomencl}
\makenomenclature

\usepackage{graphicx}
\graphicspath{ {Figures/} }

\usepackage{amsmath,bm}
\usepackage{amssymb}

\usepackage{cite}
\usepackage{amsmath}
\usepackage{listings}
\usepackage{color}
\definecolor{mygreen}{rgb}{0,0.6,0}
\definecolor{mygray}{rgb}{0.5,0.5,0.5}
\definecolor{mymauve}{rgb}{0.58,0,0.82}
\usepackage{algpseudocode}

\usepackage{array}

\ifCLASSOPTIONcompsoc
  \usepackage[caption=false,font=normalsize,labelfont=sf,textfont=sf]{subfig}
\else
  \usepackage[caption=false,font=footnotesize]{subfig}
\fi

\usepackage{stfloats}
\usepackage{url}

\begin{document}

%
\title{Full Workspace Generation of Serial-link Manipulators by Deep Learning based Jacobian Estimation}

\nomenclature[04]{$\fontfamily{phv}\selectfont \textbf{A}$}{A tensor}%
\nomenclature[06]{$\textbf{A}^T$}{Transpose of matrix $\textbf{A}$}%
\nomenclature[07]{det$(\textbf{A})$}{Determinant of $\textbf{A}$}%


\nomenclature[11]{$p($\textbf{x}$)$}{A probability distribution over a continuous variable}%
\nomenclature[12]{x $\sim p$}{Random variable $x$ has distribution $p$}%
\nomenclature[13]{$\mathbb{E}_{x \sim p}[f(x)]$}{Expectation of $f(x)$ with respect to $P(x)$}%
\nomenclature[14]{$\mathcal{N}(\textbf{\textit{x}};\mu,\sigma)$}{Gaussian distribution over $\textit{\textbf{x}}$ with mean $\mu$ and
covariance $\sigma$ }%

\nomenclature[15]{$\begin{Bmatrix} a,b,c \end{Bmatrix}$ or $\mathbb{F}$}{A set}%
\nomenclature[19]{$\mathbb{B}$}{Boolean domain}%
\nomenclature[21]{$SO(n)$}{Special orthogonal group, the set of all orientations in $n$ dimensions}%
\nomenclature[22]{$\textbf{\textit{R}}$}{Orthogonal rotation matrix, $\textbf{\textit{R}} \in SO(3)$}%
\nomenclature[23]{$\mathcal{G}$}{A Graph}%
\nomenclature[24]{$Pa_{\mathcal{G}}(x_i)$}{The parents of $x_i$ in $\mathcal{G}$}%

\nomenclature[25]{$f(\textbf{\textit{x}};\theta)$}{A function of $\textbf{\textit{x}}$ parametrized by $\theta$}%
\nomenclature[26]{$\textbf{H}(f)(\textbf{\textit{x}})$}{The Hessian matrix of $f$ at input point $\textbf{\textit{x}}$}%
\nomenclature[27]{$\xi$}{Abstract representation of 3-dimensional Cartesian pose}%
\nomenclature[28]{$\xi (\textbf{\textit{a}},\textbf{\textit{R}})$}{Pose construction function that takes in a position vector and an orientation matrix and returns a pose in $SE(3)$}%
\nomenclature[29]{$\mathcal{K}(\textbf{q})$}{Forward kinematic function of a serial-link robot}%
\nomenclature[30]{$\mathcal{K}^{-1}(\xi)$}{Inverse kinematics}%
\nomenclature[32]{$\textbf{\textit{x}}^{(i)}$}{The $i$-th example from a dataset}%
\nomenclature[33]{$\textbf{\textit{y}}^{(i)}$}{The target associated with $\textbf{\textit{x}}^{(i)}$ for supervised learning}%
\nomenclature[34]{$||\textbf{\textit{x}}||$}{$L^2$ norm of $\textbf{\textit{x}}$}%
\nomenclature[35]{$\delta($x$;\mu)$ or $\delta(\mu)$}{Dirac delta distribution of a scalar random variable centered at $\mu$}%
\nomenclature[36]{$c$}{Constant with significance indicated by context}%
\nomenclature[37]{$\mathcal{U}(a_{min},a_{max})$}{Uniform distribution of a random variable on the interval $[a_{min},a_{max}]$}%

\author{\IEEEauthorblockN{Peiyuan Liao}
\IEEEauthorblockA{
Kent Artificial Intelligence Laboratory\\
Kent, Connecticut 06757\\
Email: liaop20@kent-school.edu}

\and
\IEEEauthorblockN{Jiajun Mao}
\IEEEauthorblockA{
Kent Artificial Intelligence Laboratory\\
Kent, Connecticut 06757\\
Email: maoj19@kent-school.edu}
}


%


\maketitle

\begin{abstract}
Apart from solving complicated problems that require a certain level of intelligence, fine-tuned deep neural networks can also create fast algorithms for slow, numerical tasks. In this paper, we introduce an improved version of \cite{Liao}'s work, a fast, deep-learning framework capable of generating the full workspace of serial-link manipulators. The architecture consists of two neural networks: an estimation net that approximates the manipulator Jacobian, and a confidence net that measures the confidence of the approximation. We also introduce M3 (Manipulability Maps of Manipulators), a MATLAB robotics library based on \cite{CorkeRobotics}(RTB), the datasets generated by which are used by this work. Results have shown that not only are the neural networks significantly faster than numerical inverse kinematics, it also offers superior accuracy when compared to other machine learning alternatives. Implementations of the algorithm (based on Keras \cite{Keras}), including benchmark evaluation script, are available at \url{https://github.com/liaopeiyuan/Jacobian-Estimation}. The M3 Library APIs and datasets are also available at \url{https://github.com/liaopeiyuan/M3}.
\end{abstract}


%
\IEEEpeerreviewmaketitle

\section{Introduction}
The algorithm of finding manipulator workspace, the set of obtainable 6-D poses given a fixed range of reachable parameters, is a well-studied subject, utilizing various techniques in rigid body kinematics and dynamics. According to the literature \cite{AW14CeccarelliEclipse,AW15LeeYoung,AW1BinaryMapCastelli,AW2WorkSpaceSE3Jin,AW3MonteCarloWorkspacePeidro,AW4WorkspaceParallelAnnTanase,AW5WorkspaceParAnnGenKuzeci,AW6WorkspaceParaCampean,AW7WorkspaceParaAlp,AW8WorkspaceParWang,AW9RobotWorkspaceNumCao,AW10ParallelWrkspcOptHerrero2015,AW11SingMapParallelMacho,AW12HybridRobotModelPisla2013,AW13NonConvexWrkspcParallelHay2002}. , The notion of workspace can be further developed into the following subdivisions:

\begin{itemize}
    
    \item Reachable Workspace: The set of points end effector could reach with at least one orientation
    
    \item Total Orientation Workspace: The set of points end effector could reach with all orientation angles in a given range
    
    \item Dexterous Workspace: The set of points end effector could reach with all orientation angles
    
    \item Orientation Workspace: The set of orientations that result in a fixed end effector location
    
    \item Constant orientation Workspace: The set of points end effector could reach with a specific orientation
    
\end{itemize}

Thus, a central idea from the current approaches is the controlling of variables: since displaying $\mathbb{R}^{3}\times SO(3)$ is not possible, constraints are added to the $SO(3)$ part so that the graph of $\mathbb{R}^3$ can be drawn \cite{ParallelMerlet}. However, as pointed out by \cite{Liao}, a prevalent issue of the current algorithms is that they require specific software platforms and have bad time complexities. An analysis done in the same work remarked that a MATLAB implementation yields a pathological $O(n^{36})$ complexity with numerical inverse kinematics. Thus, a natural solution comes from the field of pattern recognition and machine learning, where the problem of manipulator workspace could be reframed into a supervised learning problem, and thus be learned by a deep neural network. In this work, we propose such a framework, seen as an improvement of the Subspace Learning algorithm in \cite{Liao}, where the full workspace of a serial-link manipulator can be generated from approximating its Jacobian matrix, if exists, at a given pose. 

\section{Background Information}

\subsection{Manipulator Jacobian}
The relationship between the change in joint angles and the change in the orientation and the coordinates of the end-effector is locally linear, which means that the spatial velocity of the end-effector can be approximated by applying a linear transformation to the rate of change of the joint angles. Consider a basic homogeneous transformation representation of a pose $\xi$. A first-order difference relationship exists between the angles that define the pose and the pose itself: \cite{CorkeRobotics}

\begin{equation}\label{poseDerivation}
\begin{split}
\frac{d\xi}{d\textit{\textbf{q}}} & \approx \frac{\xi( \textit{\textbf{q}} + \Delta \textit{\textbf{q}}) - \xi(\textit{\textbf{q}})}{\Delta \textit{\textbf{q}}}
\\ & \approx \frac{1}{\Delta \textit{\textbf{q}}}\begin{bmatrix}
R(\textit{\textbf{q}} + \Delta \textit{\textbf{q}}) - R(\textit{\textbf{q}}) & \begin{matrix}
\Delta x \\
\Delta y \\
\Delta z 
\end{matrix} \\
0_{1 \times 3} & 0
\end{bmatrix}
\end{split}
\end{equation}

\begin{figure}
    \centering
    \includegraphics[width=.5\textwidth]{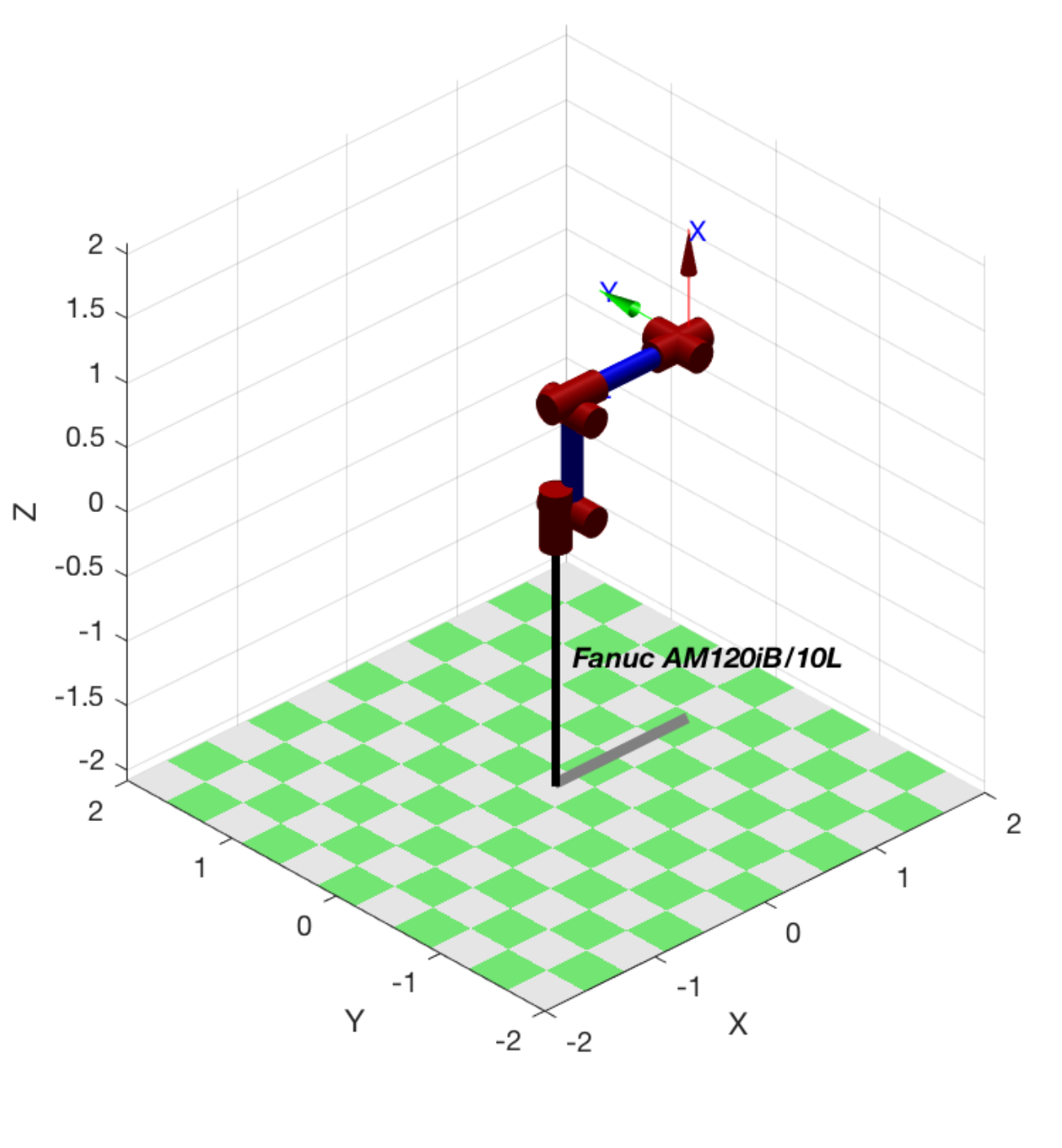}
    \caption{Fanuc Am120iB/10L}
    \label{fig:mesh9}
\end{figure}

The first part of the analysis is trying to transform the upper right part into the linear velocity of the end-effector. In this paper, a specific example is examined where the numerical linear velocity at a specific joint angle is calculated. Figure \eqref{fig:mesh9} shows the Fanuc AM120iB/10L, an industrial 6-DOF serial-link manipulator created in  \cite{CorkeRobotics} using standard Denvait-Hartenberg parameters. The robot's pose is currently $ \xi \in SE(3) = \begin{bmatrix}
1 & 0 & 0 & 1.02 \\
0 & 1 & 0 & 0 \\
0 & 0 & 1 & -1.06 \\
0 & 0 & 0 & 1
\end{bmatrix} $, with joint coordinates of $\begin{bmatrix}
0 & 0 &0 &0 &0 &0 
\end{bmatrix}$ .

Now, consider moving the first joint of the robot by an infinitely small angle $dq$, approximated by the value $10^{-9}$. Applying Equation \eqref{poseDerivation} results in

\begin{equation}\label{positionDerivation}
    \begin{split}
\frac{\partial \xi}{\partial q_1} & \rvert { {\textit{\textbf{q}}  = \begin{bmatrix}
0,0,0,0,0,0 
    \end{bmatrix}} } \\ & \approx \frac {\mathcal{K}( \begin{bmatrix}
    0_{1 \times 6}
    \end{bmatrix} + \begin{bmatrix}
    10^{-9} & 0_{1 \times 5}
    \end{bmatrix} ) - \mathcal{K}(\begin{bmatrix}
    0_{1 \times 6}
    \end{bmatrix} ) } {10^{-9}} \\
                  & \approx \begin{bmatrix}
                        0 & -1 & 0 & 0 \\
                        1 & 1 & 0 & 1.02 \\
                        0 & 0 & 0 & 0 \\
                        0 & 0 & 0 & 0
                    \end{bmatrix}    
    \end{split}
\end{equation}

Equating elements from this matrix to those in Equation \eqref{poseDerivation} gives

\begin{equation}\label{posDe2}
\begin{bmatrix}
\Delta x \\
\Delta y \\
\Delta z
\end{bmatrix} = \begin{bmatrix}
0 \\
1.02 \\
0 \\
\end{bmatrix}\Delta q_1
\end{equation} 

Finally, since velocity is the relationship between distance and time, dividing both sides by an infinitely small change of time $\Delta t$ gives

\begin{equation}\label{posDe3}
\begin{bmatrix}
\dot{x} \\
\dot{y} \\
\dot{z} 
\end{bmatrix} = \begin{bmatrix}
0 \\
1.02 \\
0 \\
\end{bmatrix}\dot{q_1}
\end{equation} 

, where $\dot{x}$ stands for the derivative of $x$ with respect to time. This conversion from the velocity of a single joint to the linear speed of the end-effector is explainable because as a joint that only rotates on a 2-D plane and has the initial position parallel to the x-axis, at the very first instant change in its angle will only result in a rightward end-effector motion.

This process is repeatable for all joints, and once added together, it demonstrates how each of the joints is affecting end-effector linear velocity.

\begin{equation}\label{posDe4}
    \begin{split}
        & \frac{d\xi}{d\textit{\textbf{q}}} \rvert { {\textit{\textbf{q}}  = \begin{bmatrix}
                0 , 0 ,0 ,0 ,0 ,0 
                \end{bmatrix}} } \\ & = \frac{\partial \xi}{\partial q_1} + \frac{\partial \xi}{\partial q_2} + \dots + \frac{\partial \xi}{\partial q_6} \\
                      & = \begin{bmatrix}
                      R(\textit{\textbf{q}} + \Delta \textit{\textbf{q}}) - R(\textit{\textbf{q}}) & \begin{matrix}
                      -1.06q_2 + 1.06q_3 + 0.1q_5 \\
                      1.02q_1 \\
                      -0.87q_2+0.1q_3 
                      \end{matrix} \\
                      0_{1 \times 3} & 0
                      \end{bmatrix}
    \end{split}
\end{equation}

This implies

\begin{equation}\label{posDe5}
\begin{bmatrix}
\dot{x} \\
\dot{y} \\
\dot{z} 
\end{bmatrix} = \begin{bmatrix}
-1.06\dot{q_2} + 1.06\dot{q_3} + 0.1\dot{q_5} \\
1.02\dot{q_1} \\
-0.87\dot{q_2}+0.1\dot{q_3}
\end{bmatrix}
\end{equation}

Although the analytic relationship between the joint angle velocity and end-effector linear velocity is still unknown, from equation \eqref{posDe5} we can know that at least when the arm just starts moving, the velocity of joint 1,2,3 and 5 dictates the movement of the end-effector. \par

For angular velocity relationship, consider the sub-matrix on the upper left corner in equation \eqref{poseDerivation}. Differentiating it results in

\begin{equation}\label{rotDe2}
\begin{split}
& \frac{\partial \textit{\textbf{R}}}{\partial q_1}\rvert { {\textit{\textbf{q}}  = \begin{bmatrix}
        0 , 0 ,0 ,0 ,0 ,0 
        \end{bmatrix}} } \\ & \approx \frac{ R(\textit{\textbf{q}} + \Delta \textit{\textbf{q}}) - R(\textit{\textbf{q}}) }{\Delta \textit{\textbf{q}}} \dot{q_1} \\
     S(\bm{\mathit{\omega}})\textit{\textbf{R}} & \approx \frac{ R(\textit{\textbf{q}} + \Delta \textit{\textbf{q}}) - R(\textit{\textbf{q}}) }{\Delta \textit{\textbf{q}}} \dot{q_1} \\
    S(\bm{\mathit{\omega}}) & \approx \frac{ R(\textit{\textbf{q}} + \Delta \textit{\textbf{q}}) - R(\textit{\textbf{q}}) }{\Delta \textit{\textbf{q}}} \textit{\textbf{R}}^{T} \dot{q_1} \\
    \omega & \approx \text{vex} \begin{pmatrix}
        \frac{ R(\textit{\textbf{q}} + \Delta \textit{\textbf{q}}) - R(\textit{\textbf{q}}) }{\Delta \textit{\textbf{q}}} \textit{\textbf{R}}^{T}
    \end{pmatrix} \dot{q_1} \\
    \begin{bmatrix}
    \omega_x \\
    \omega_y \\
    \omega_z
    \end{bmatrix} & = \begin{bmatrix}
    0 \\ 0 \\ 1
    \end{bmatrix} \dot{q_1}
\end{split}
\end{equation}

,where $\text{vex}(\textit{\textbf{S}})$ is the inverse operation of the skew-symmetric matrix.


Finally, after establishing relationships between the joint angle velocities and the translational velocity angular velocities separately, we can arrive at a cumulative representation of this relationship - the manipulator Jacobian. Taking the derivative of the forward kinematics function, we obtain 

\begin{equation}\label{jacobian}
\begin{split}
    \frac{d \xi}{d\textit{\textbf{q}}} & = \frac{d}{d\textit{\textbf{q}}} \mathcal{K}(\textit{\textbf{q}})\\
    \bm{\mathit{\nu}} & = J(\textit{\textbf{q}})\dot{\textit{\textbf{q}}}
\end{split}
\end{equation}

,where

\begin{equation*}
\begin{split}
    \bm{\mathit{\nu}} & = \begin{bmatrix} v_x & v_y & v_z & \omega_x & \omega_y & \omega_z \end{bmatrix} \\
    J(\textbf{\textit{q}}) & \in \mathbb{R}^{6 \times N}
\end{split}
\end{equation*}  
,$N$ being the degree-of-freedom of the robot.

\subsection{Workspace Mapping Seen as a Supervised Learning Problem}
According to \cite{GoodfellowDeep}, a supervised learning problem can be seen as finding a random vector $\textbf{x}$'s corresponding value $\textbf{y}$ by estimating $p(\textbf{y}|\textbf{x})$. This work uses the following steps to transform a workspace mapping problem to a supervised learning problem \cite{Liao}:

\begin{itemize}
    
    \item For a given all-revolute, serial-link manipulator, we first have a left-invertible forward kinematics function:
\begin{equation}\label{fkine}
\begin{split}     
     \mathcal{K}: \mathbb{S}^n \rightarrow \mathbb{R}^3 \times SO(3)
\end{split}
\end{equation}

     , where $\mathbb{S}$ is the set of all angles in the interval $[0, 2\pi)$. This maps a point in the parameter space of the manipulator to a pose in 6-D space. In an object-oriented programming perspective, for a \verb| Manipulator |  object we would have two methods: \verb| Manipulator.forwardKine(q)| that corresponds to $\mathcal{K}(\textit{\textbf{q}})$  and \verb| Manipulator.inverseKine(xi)| that corresponds to $\mathcal{K}^{-1}(\xi)$ (we use $^{-1}$ here to denote left inverse). This may be computed numerically of analytically, and, for quick reference, see the Appendices in \cite{Liao}. For a closer look into the details, one may refer to \cite{CorkeRobotics} or \cite{SicilianoRobotics}.
    \item We then define the Jacobian mapping function 
    \begin{equation}\label{jacobmap}
\begin{split}    
f: (\mathbb{R}^{m})^{n}\times \mathbb{R}^3 \times SO(3) \rightarrow  \mathbb{R}^{6 \times n}
\end{split}
\end{equation}
that maps a manipulator with a corresponding pose to it's corresponding Jacobian matrix, where $n$ is the number of links and $m$ is the number of parameters needed to describe a single link. \cite{Liao} provides a specific from of the above function, but this expression aim to provide a general form.
    
    \item However, remark that $f$ is ill-defined because $\mathcal{K}$ is not right-invertible-a manipulator cannot reach infinite space-and thus $f$ is not left-total.
    
    \item To resolve this issue, we introduce an improved version of $f$ :
    
\begin{equation}\label{improvedJacob}
\begin{split}     
     f_1 &: (\mathbb{R}^{m})^{n}\times \mathbb{R}^3 \times SO(3) \rightarrow  \begin{Bmatrix}0,1\end{Bmatrix} \\
     f_2 &: (\mathbb{R}^{m})^{n}\times \begin{Bmatrix}0,1\end{Bmatrix} \rightarrow  \mathbb{R}^{6 \times n}
\end{split}
\end{equation}
    They are defined via: 
    \begin{enumerate}
    \item $f_1(\textbf{m},\xi) = 1$ if $\exists \textbf{q} \in \mathbb{S}^n$ such that $\mathcal{K}(q)=\xi$, denoted $P(\textbf{m},\xi)$; $f_1(\textbf{m},\xi) = 0$ if $\neg P(\textbf{m},\xi)$.
    \item $f_2(\textbf{m},b) = \inf_{6 \times n}$ if $b=0$; otherwise,  $f_2(\textbf{m},f_1(\textbf{m},\xi)) = f(\textbf{m},\xi)$.
    \end{enumerate}
        
    \item Design $\textbf{x}$ on a implementation basis: range of scopes that corresponds to $p(\xi)$ and the "manipulator distribution" $p(\textbf{m})$, the latter usually uniform. By sampling from  $p(\textbf{x})$ and evaluating $f_2(\textbf{m},f_1(\textbf{m},\xi))$ numerically, we have created a supervised learning problem.    
\end{itemize}
Implementation-wise, according to empirical-risk minimization \cite{GoodfellowDeep},

\begin{equation}
    \mathbb{E}_{\textbf{x},\textbf{y}\sim \widehat{p}_{data} (\textbf{x},\textbf{y})} [L(f(\textbf{x};\mathbf{\theta}),\textbf{y})] = \frac{1}{m}\sum^{m}_{i=1} L(f(\textbf{x}^{(i)}; \mathbf{\theta}),\textbf{y}^{(i)})
\end{equation}

The above can be done in a discrete manner, where a dataset can be created by sampling from $\textbf{x} \sim p(\textbf{x})$, and numerically compute the corresponding values of each sample, without having to worry about the analytic form of $p(\textbf{y})$. 

\section{Related Works}

There are a number of attempts to apply deep learning to problems in manipulator robotics, either to obtain the workspace of robots or to achieve other goals in general. An obvious application is the inverse kinematics itself, due to its non-linearity and the NP-completeness of optimal Jacobian accumulation \cite{NPJacobian}. While early approaches often focus on a particular model with a tailored network architecture \cite{IkineMLPGuez, IkineDEFAnetDaunicht, PathControlIkineAnnLou}, more recent studies begin to generalize so that the deep network can solve the problem for a broader spectrum of kinematic structures \cite{Ikine6rBingul, IkineGenKoker, IkineAnnFeng}. Another fascinating field is the task-learning of humanoid robots, where their arms are seen as kinematically redundant manipulators \cite{FeatFaceRobotManiFinn}. The goal is to learn basic arm actions described in visual state representation from video feeds rather than the pose of the desired object. The anthropomorphic nature of the problem is intriguing, and by utilizing the spatial encoders, the tested robot is capable of performing actions such as scooping a bag of rice with a spatula.

\subsection{Improvements over Subspace Learning}
As a follow-up study of \cite{Liao}, this paper obtains incremental improvements over several aspects:

\begin{enumerate}

    \item Not only predicting the existence of inverse kinematics, but the new framework also offers an approximation of the Jacobian matrix at a given pose, which is useful for a spectrum of tasks such as resolved rate motion control \cite{CorkeRobotics} and manipulability estimation \cite{manipulability}. 
    
    \item A Python implementation is provided with pretrained weights, making it much more portable than the previous MATLAB implementation.
    
    \item More advanced neural architectures are used, including new activation functions \cite{PReLU}, batch normalization \cite{batchnorm}, and dropout \cite{dropout} that helps the model to generalize better.
    
    \item Newer gradient descent optimizers are tested on the framework, like RMSprop \cite{RMSprop}, Adagrad \cite{Adagrad}, Adam \cite{Adam,convAdam}, Nadam \cite{Nadam}, Adadelta \cite{Adadelta} and Adamax \cite{Adam}, all provided by Keras \cite{Keras}. 
\end{enumerate}

\section{M3 Library and Dataset}
Along with the JacobianNet framework, we also release the Manipulability Map for Manipulators (M3) library and datasets, a MATLAB robotics library designed for deep learning related sampling of manipulator parameters and its corresponding outputs. The library functions are based on \cite{Liao}'s released codes, with improvements in readability and extra features.

The main function in the library's API is \verb|NNsample.m|, with the following signature and options:

\begin{lstlisting}[language=octave]
function [features,labels]=NNsample(num,parallel,varargin)
    opt.format='csv';
    opt.variant='CGAN';
    opt.mani='kine';
    opt.DOF=6;
    opt.type='spherical';
    opt.r=0.5;
    opt.dist='uniform';
    opt.poseR=0.8;
    opt.ikine='analytic';
    opt.plim=50;
    opt.testRatio=0.01;
    opt.cutoff=0.03;
\end{lstlisting}

Currently, the library supports dataset generation for generative modeling, inverse kinematics, and dynamics studies. This can be tuned by setting \verb|opt.variant| to different values. \verb|parallel| argument requires a boolean input, which determines if parallel pools would be used to accelerate computing. There are also other scripts in the library, some used by \verb|NNsample.m|, which might also be useful on their own: 

\begin{itemize}
 \item \verb|cell2csv.m| \cite{cell2csv}, which turns a cell array into a \verb|.csv| file. 
 \item \verb|threshold.m|, used mainly for generative modeling research, which determines if a sample is considered as belonging to the true distribution.
 \item \verb|randPose.m|, which returns a random pose given scope parameters.
 \item \verb|genWorkspace.m|, a canonical implementation of the discretized workspace algorithm (both for constant orientation workspace and orientation workspace)
 \item \verb|randDyna.m| and \verb|randKine.m|, which returns a randomly sampled manipulator object with its vectorized representation.
\end{itemize}

All codes are available at \url{https://github.com/liaopeiyuan/M3} published under the MIT License,  maintained by the Kent Artificial Intelligence Laboratory.

The dataset accompanied by the library, which is used to train the network proposed above, is available along with the models at \url{https://github.com/liaopeiyuan/Jacobian-Estimation}, and consists of the following files:

\begin{itemize}
 \item \verb|conf_feature_test.csv|
 \item \verb|conf_feature_train.csv|
 \item \verb|conf_label_test.csv|
 \item \verb|conf_label_train.csv|
 \item \verb|conf_feature_dyna_test.csv|
 \item \verb|conf_feature_dyna_train.csv|
 \item \verb|conf_label_dyna_test.csv|
 \item \verb|conf_label_dyna_train.csv|
 \item \verb|jacob_feature_test.csv|
 \item \verb|jacob_feature_train.csv| 
 \item \verb|jacob_label_test.csv|
 \item \verb|jacob_label_train.csv| 
 \item \verb|jacob0_feature_test.csv|
 \item \verb|jacob0_feature_train.csv|
 \item \verb|jacob0_label_test.csv|
 \item \verb|jacob0_label_train.csv| 
\end{itemize}

Files with \verb|train| prefix are used for train/validation split, and \verb|test| prefix for benchmarking results. Remark that files with \verb|conf| prefix are used to train the confidence net, and files with \verb|jacob| prefix are used to train the confidence nets. Files with \verb|jacob| prefix record the Jacobian matrix in the end-effector frame while files with \verb|jacob0| prefix record the Jacobian matrix in the world frame.

\begin{table}[h]
\caption{Preliminary Data Analysis: Part 1} 
\centering 
\begin{tabular}{c rrrrrrr} 
    \hline\hline 
    File Name& \# Targets & \# Features & \# Samples & Regres./Class. \\ [0.5ex]
\hline 
\verb|conf_*_test| & 1 & 24 & 3000 & Classification \\
\verb|conf_*_train| & 1 & 24 & 297000 & Classification \\

  \verb|conf_*_dyna_test| & 1 & 96 & 2000 & Classification\\
  \verb|conf_*_dyna_train| & 1 & 96 & 198000 & Classification\\

  \verb|jacob_*_test| & 36 & 96 & 2000 & Regression\\
  \verb|jacob_*_train| & 36 & 96 & 198000 & Regression\\

  \verb|jacob0_*_test| & 36 & 96 & 2000 & Regression\\
  \verb|jacob0_*_train| & 36 & 96 & 198000 & Regression\\[1ex] 

\hline 
\end{tabular}
\label{tab:hresult}
\end{table}

\begin{figure}
    \centering
    \includegraphics[width=.5\textwidth]{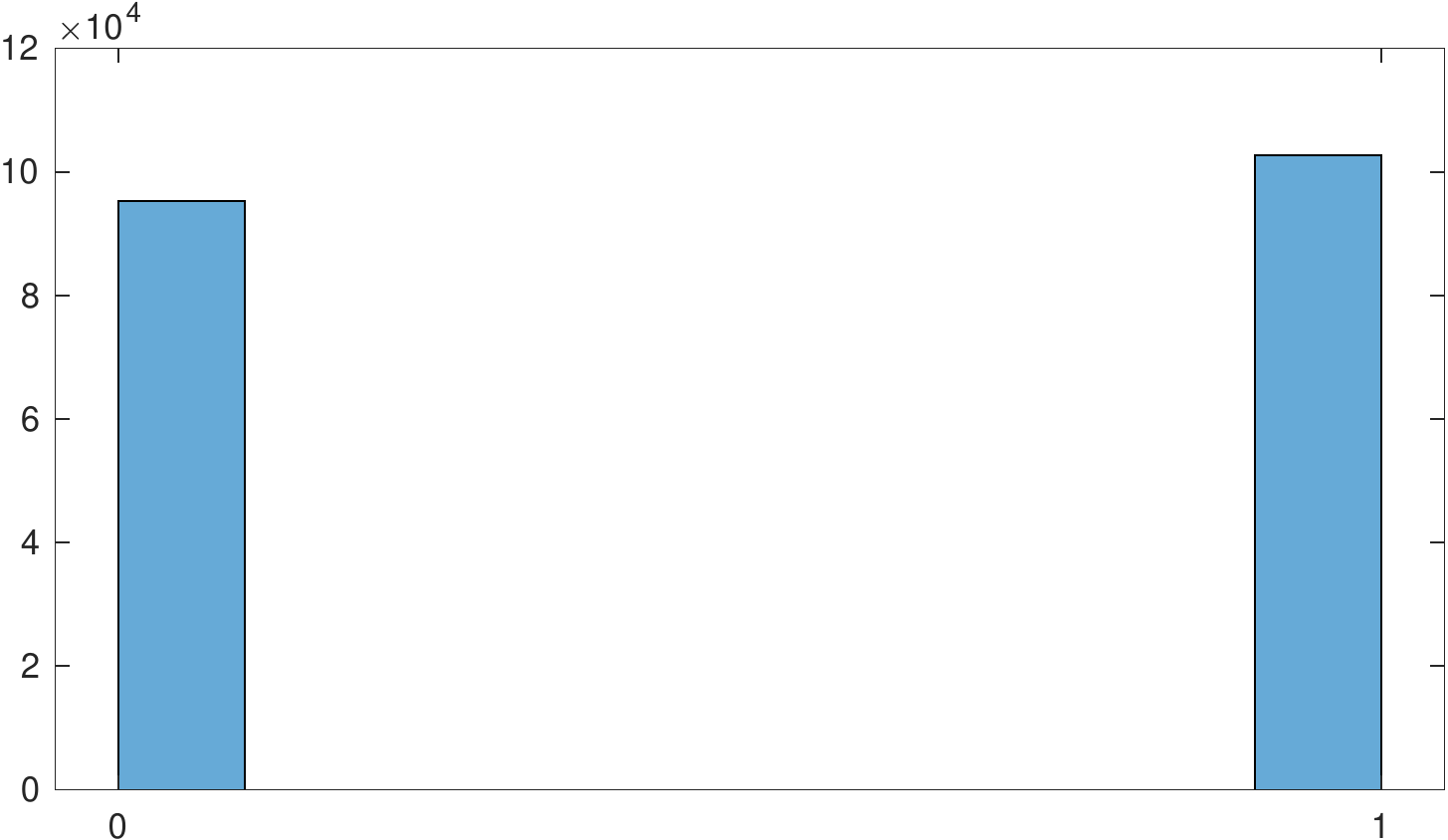}
    \caption{Ratio of positive samples versus negatives in the confidence net training samples, dynamics version}
    \label{fig:mesh10}
\end{figure}

\begin{figure}
    \centering
    \includegraphics[width=.5\textwidth]{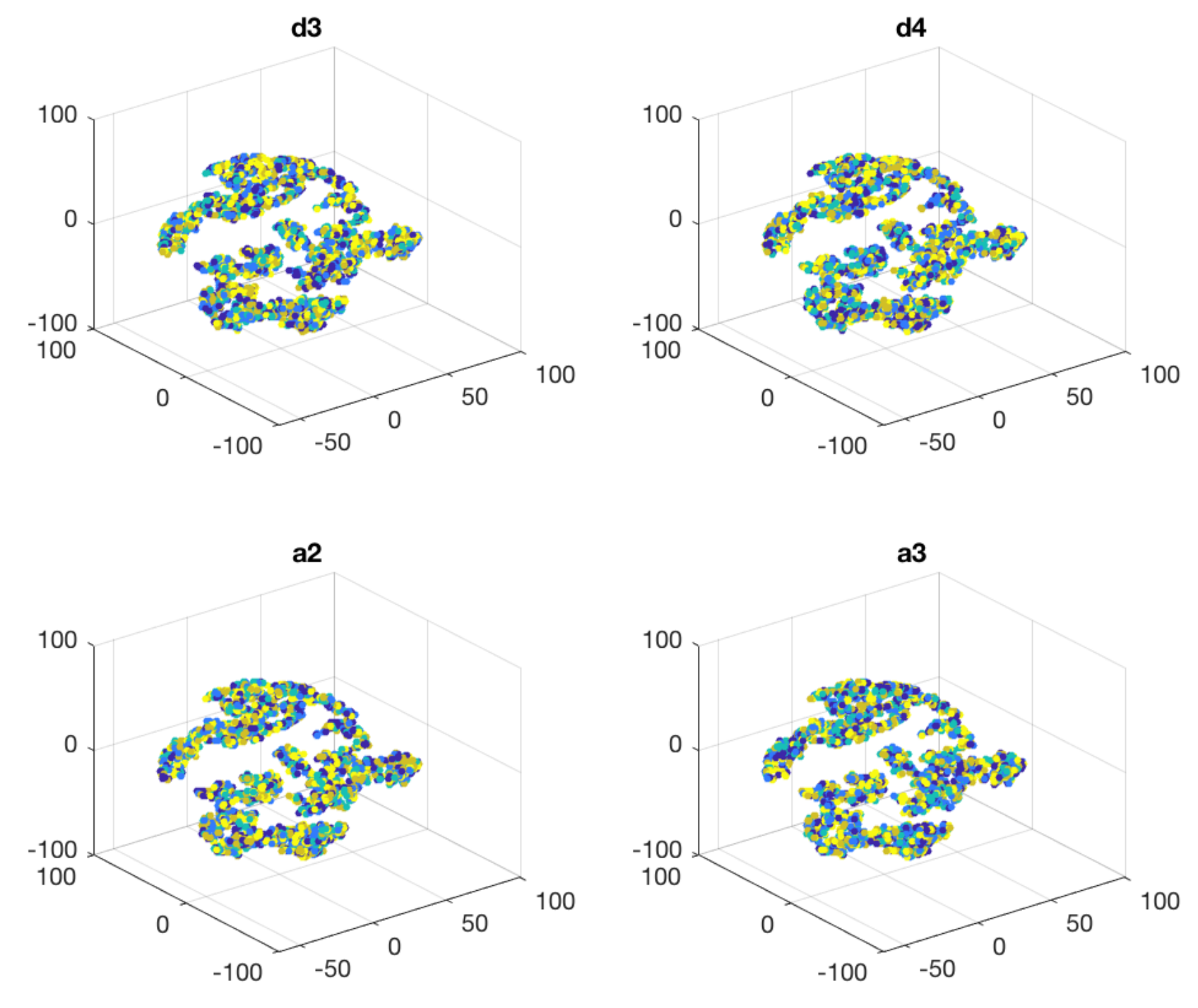}
    \caption{t-SNE plot of test input features for confidence net, stratified by $d3$, $d4$, $a2$, $a3$ respectively}
    \label{fig:mesh10}
\end{figure}

\begin{table}[h]
\caption{Preliminary Data Analysis: Part 2 (features)} 
\centering 
\begin{tabular}{c rrr} 
    \hline\hline 
    File Name& 2,3,7,8 (DH params) & Constant Cols. \\ [0.5ex]
\hline 
\verb|conf_*_test|  &$[0,0.5]$ & $\begin{Bmatrix}0,1,4\sim6,9\sim17\end{Bmatrix}$\\
\verb|conf_*_train|  &$[0,0.5]$ & $\begin{Bmatrix}0,1,4\sim6,9\sim17\end{Bmatrix}$\\

  \verb|conf_*_dyna_test| &$[0,0.5]$ & $\begin{Bmatrix}0,1,4\sim6,9\sim17\end{Bmatrix}$\\
  \verb|conf_*_dyna_train| &$[0,0.5]$ & $\begin{Bmatrix}0,1,4\sim6,9\sim17\end{Bmatrix}$\\

  \verb|jacob_*_test| &$[0,0.5]$ & $\begin{Bmatrix}0,1,4\sim6,9\sim17\end{Bmatrix}$\\
  \verb|jacob_*_train| &$[0,0.5]$ & $\begin{Bmatrix}0,1,4\sim6,9\sim17\end{Bmatrix}$ \\

  \verb|jacob0_*_test| &$[0,0.5]$& $\begin{Bmatrix}0,1,4\sim6,9\sim17\end{Bmatrix}$\\
  \verb|jacob0_*_train| &$[0,0.5]$& $\begin{Bmatrix}0,1,4\sim6,9\sim17\end{Bmatrix}$\\[1ex] 

\hline 
\end{tabular}
\label{tab:hresult2}
\end{table}

\begin{table}[h]
\caption{Preliminary Data Analysis: Part 3 (feature column range)} 
\centering 
\begin{tabular}{c rrrrr} 
    \hline\hline 
    File Name& 18$\sim$23 (m) & 24$\sim$41 (r) & 42$\sim$59 (I) \\ [0.5ex]
\hline 
  \verb|conf_*_dyna_test| &$[0,10]$ &$[-0.05,0.05]$ &$[0,1]$ \\
  \verb|conf_*_dyna_train| &$[0,10]$ &$[-0.05,0.05]$ &$[0,1]$\\

  \verb|jacob_*_test| &$[0,10]$ &$[-0.05,0.05]$ &$[0,1]$ \\
  \verb|jacob_*_train| &$[0,10]$ &$[-0.05,0.05]$ &$[0,1]$ \\

  \verb|jacob0_*_test| &$[0,10]$&$[-0.05,0.05]$&$[0,1]$  \\
  \verb|jacob0_*_train| &$[0,10]$&$[-0.05,0.05]$ &$[0,1]$\\[1ex] 

\hline 
\end{tabular}
\label{tab:hresult3}
\end{table}

\begin{table}[h]
\caption{Preliminary Data Analysis: Part 4 (feature column range)} 
\centering 
\begin{tabular}{c rrrrr} 
    \hline\hline 
    File Name& 60$\sim$65 (B) & \multicolumn{2}{c}{66$\sim$77 (Tc)} & 78$\sim$83 (G) \\ [0.5ex]
\hline 
  \verb|conf_*_dyna_test| &$[0,0.005]$ &$[0,0.5]$ &$[-0.5,0]$&$[-50,50]$ \\
  \verb|conf_*_dyna_train| &$[0,0.005]$ &$[0,0.5]$ &$[-0.5,0]$&$[-50,50]$\\

  \verb|jacob_*_test| &$[0,0.005]$ &$[0,0.5]$ &$[-0.5,0]$ &$[-50,50]$\\
  \verb|jacob_*_train| &$[0,0.005]$ &$[0,0.5]$ &$[-0.5,0]$&$[-50,50]$ \\

  \verb|jacob0_*_test| &$[0,0.005]$&$[0,0.5]$&$[-0.5,0]$ &$[-50,50]$ \\
  \verb|jacob0_*_train| &$[0,0.005]$&$[0,0.5]$ &$[-0.5,0]$&$[-50,50]$\\[1ex] 

\hline 
\end{tabular}
\label{tab:hresult4}
\end{table}

\begin{table}[h]
\caption{Preliminary Data Analysis: Part 5 (feature column range)} 
\centering 
\begin{tabular}{c rrrrr} 
    \hline\hline 
    File Name& 84$\sim$89 (Jm) & 90$\sim$92 ($\mathbb{R}^3$) & 93$\sim$95 ($\mathbb{S}^3$)  \\ [0.5ex]
\hline 
  \verb|conf_*_dyna_test| &$[0,5\times10^{-4}]$ &$[-0.4,0.4]$ &$[0,2\pi]$ \\
  \verb|conf_*_dyna_train| &$[0,5\times10^{-4}]$ &$[-0.4,0.4]$ &$[0,2\pi]$\\

  \verb|jacob_*_test| &$[0,5\times10^{-4}]$ &$[-0.4,0.4]$ &$[0,2\pi]$ \\
  \verb|jacob_*_train| &$[0,5\times10^{-4}]$ &$[-0.4,0.4]$ &$[0,2\pi]$ \\

  \verb|jacob0_*_test| &$[0,5\times10^{-4}]$&$[-0.4,0.4]$&$[0,2\pi]$  \\
  \verb|jacob0_*_train| &$[0,5\times10^{-4}]$&$[-0.4,0.4]$ &$[0,2\pi]$\\[1ex] 

\hline 
\end{tabular}
\label{tab:hresult5}
\end{table}

\begin{figure}
    \centering
    \includegraphics[width=.5\textwidth]{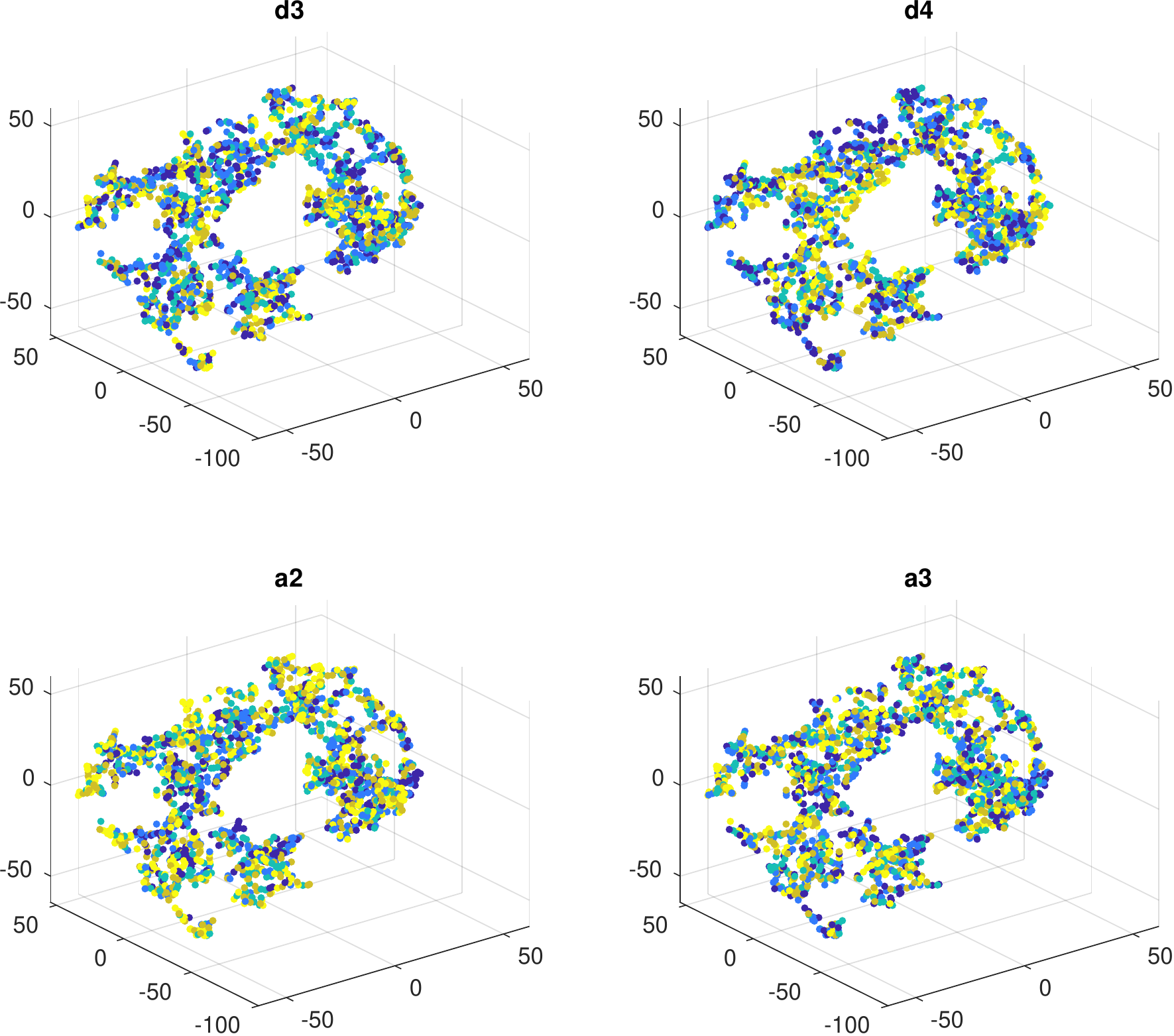}
    \caption{t-SNE plot of test Jacobian matrix (in world frame), stratified by $d3$, $d4$, $a2$, $a3$ respectively}
    \label{fig:mesh10}
\end{figure}

\section{Network Architecture}
The confidence network serves the purpose of indicating that whether a numerical solution for Jacobian matrix can be estimated from the input representation. It consists of $8$ dense layers and is compiled with binary cross entropy loss:

\begin{equation}
L_{\text{bce}}(y, \hat{y}) = -\sum_i y_i \log \hat{y}_i
\end{equation}

\begin{figure}
    \centering
    \includegraphics[width=.5\textwidth]{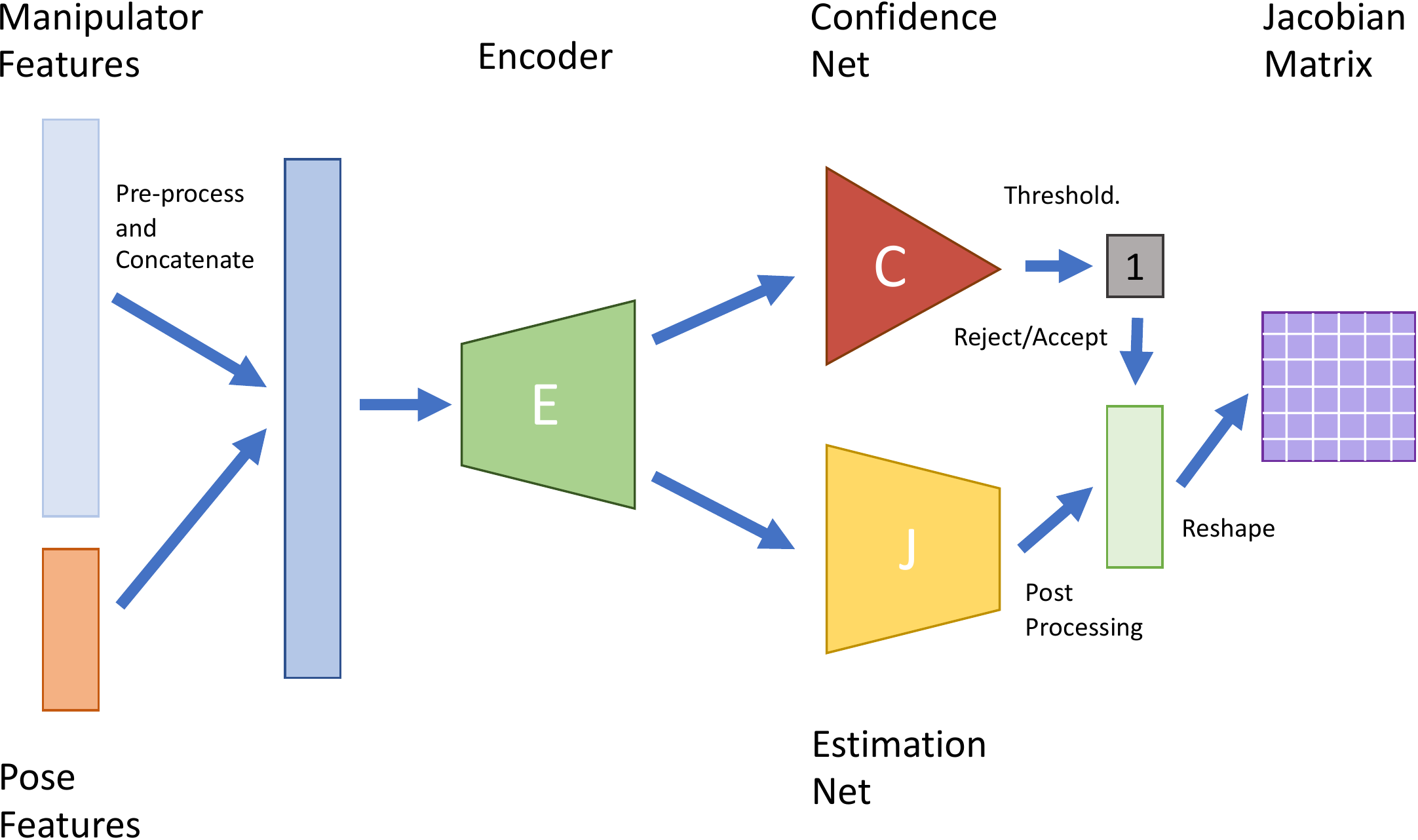}
    \caption{JacobianNet Architecture}
    \label{fig:archnew}
\end{figure}

The estimation network, as its name suggests, is the network that estimates a Jacobian matrix as the output from its input representation. It as well contains $8$ dense layers, and it outputs a $22*1$ vector that is subsequently reshaped into a complete Jacobian matrix. Since the task is treated as regression, the network is compiled with mean squared error loss:

\begin{equation}
L_{\text{mse}}(y, \hat{y}) = \frac{1}{n}\sum_{i=1}^n (y_i-\hat{y}_i)^2
\end{equation}
As suggested in Figure \ref{fig:archnew}, the confidence network, and the estimation share the same encoded representation from the encoder network. Noted that the function of the confidence network is to indicate whether a solution exists for the encoded representation, the activation of the estimation network depends on the result of the confidence network.
To elaborate, if the confidence returns the result indicating that a solution can be estimated from the encoded representation, then will the estimation network be initiated and estimate the Jacobian matrix.

As shown in Figure \ref{fig:archnew} and described above, the combined model will produce a Jacobian matrix only when the confidence network indicates that a solution is possible. This model is constructed because we believe that this model can provide us with few advantages:
\begin{enumerate}
\item \textit{Memory-Efficiency.} Since some encoded representation does not need to be estimated by the estimation net if it is considered as insolvable by the confidence net, this practice might be able to save memory from preventing it to be used to store estimation outcome that will not be able to yield a Jacobian matrix.
\item \textit{Full-Pipeline.} The combined model can offer its users the ability to obtain a Jacobian matrix directly from certain features of a robot arm, assisting them to obtain the manipulator Jacobian matrix more conveniently and speedily.
\end{enumerate}

The input for the combined model consists of a tensor concatenated from two data segment - manipulator features and pose feature, which is subsequently fed into the encoder network shown in $f_2$. The encoder network then encodes the input tensor into a $2150*1$ vector representation that will be used to estimate the resulting Jacobian matrix and the relating confidence.
Remark that the confidence net is a neural network trained to determine whether a numerical solution exists for the encoded representation of a Jacobian matrix based on a manually set threshold shown in Figure \ref{fig:archnew}. If the outcome of the confidence net is below the threshold, the representation is considered insolvable, discarded and never examined by the estimation net.
Lastly, the Jacobian estimation net, originally proposed in \cite{Liao} produces a vector from the input representation. The output vector is subsequently reshaped into the estimated Jacobian matrix representing the manipulability workspace of the robot arm. 
\subsection{Training}
Three datasets are used to train the neural networks correspondingly: the kinematics confidence dataset for confidence network, dynamics Jacobian estimation dataset for estimation network, and dynamics estimation/confidence dataset for the combined model. 
For the encoder network, batch normalization\cite{batchnorm} is used after the last four dense layers as well as the output layer with PReLU\cite{PReLU} activation function and dropout\cite{dropout} layer with a dropout rate of $0.5$.
A similar structure of batch normalization, PReLU and dropout are also applied to the first four layers of the estimation and confidence network. For the training of the combined network, Adam optimizer\cite{Adam} is employed with its default parameters ($\beta1-0.9$, $\beta2-0.999$, $\alpha-0.001$, $\epsilon-10^{-8}$), a batch size of 4096 and epoch size of 25. 
To prevent the issue of overfitting, a validation split value is set to separate the data set into the training set and validation set. For the training of all the networks, the value for the validation split is kept at $0.2$, meaning that $20\%$ of the dataset is set to be the validation set.

There are different models constructed for comparison:
\begin{enumerate}
    \item \textit{Baseline Model.} Baseline model is constructed with default Adam optimizer\cite{Adam} and it mainly serves as a point of comparison for accuracy between the method of generating manipulator Jacobian matrix proposed in this paper and other machine learning methods proposed for similar proposes. Also, the speed of generating manipulator Jacobian matrix is compared between our method that traditional kinematic methods.
    \item \textit{Comparison Models.} Comparison models are a set of methods that utilize different types of gradient descent method. We constructed this set of models to observe different gradient step methods' influence on the result of the training of the network and the general outcome/accuracy of the combined model. The detailed comparison between different models will be elaborated in later chapters.
\end{enumerate}
\subsection{Data Augmentation}
To accelerate convergence, several augmentations are employed. All of our data augmentation functions are a part of the Scikit-Learn library\cite{scikit-learn}.
When training each of the individual networks (encoder, confidence, etc.), input features are augmented through scaling to min and unit measure using the function \verb|scale|. In addition, data augmentation is applied to the output of the estimation network for target features by using min-max scaling from $-1$ to $1$. To reiterate, data augmentation is applied to the input of all three networks and the output of the estimation network aiming for faster convergence during the training process.

\section{Experiments}
\subsection{Benchmarking Results}
According to Table \ref{tab:conf} and Table \ref{tab:est}, we can see the following advantages of our proposed model:
\begin{enumerate}
    \item \textit{Greater Accuracy.} One purpose that we are seeking to achieve with the combined model proposed in this paper is greater accuracy when compared to other machine learning models aiming to solve similar problems.
    \item \textit{Greater Speed.} Another characteristic of the proposed combined model is that when comparing to traditional numerical/analytic inverse kinematics (\verb|ikine.m| / \verb|ikine6s.m| in RTB\cite{CorkeRobotics}) for solving the manipulator Jacobian; it can come up with the result in greater speed.
\end{enumerate}
Specifically, JSC stands for Jaccard similarity coefficient, MCC stands for Matthews Correlation Coefficient, 0-1 stands for the 0-1 loss, EVS stands for explained variance score, and the number in the parenthesis indicates the thresholding value of continuous output.

\begin{table}[h]
\caption{Confidence net benchmark} 
\centering 
\begin{tabular}{c rrrrrrrr} 
    \hline\hline 
    Method  & JSC & 0-1 & Prec. & Rec. & MCC & Avg.Time\\ [0.4ex]
\hline 
  NN(.5)  & 0.983 & 0.017 & 0.97 & 0.99 & 0.967 & $2*10^{-4}$s\\
  NN(.25)  & 0.98 & 0.02 & 0.96 & \textbf{1.00} & 0.961 & $2*10^{-4}$s\\
  NN(.75)  & \textbf{0.985} & \textbf{0.015} & \textbf{0.98} & 0.99 & \textbf{0.970} & $2*10^{-4}$s\\
  SVM(RBF)  & 0.930 & 0.070 & 0.92 & 0.95 & 0.861 & $5*10^{-3}$s\\
  SVM(Sigmoid)  & 0.643 & 0.357 & 0.65 & 0.67 & 0.285 & $5*10^{-3}$s\\
  SVM(Lin. L2)  & 0.751 & 0.249 & 0.76 & 0.77 & 0.502 & $5*10^{-7}$s\\
  LogisticReg.  & 0.752 & 0.248 & 0.75 & 0.77 & 0.504 & $3*10^{-7}$s\\
  RidgeReg.(0.5)  & 0.750 & 0.250 & 0.75 & 0.76 & 0.500 & $\mathbf{1*10^{-7}}$s\\
  Naive Bayes  & 0.803 & 0.197 & 0.79 & 0.83 & 0.607 & $2*10^{-7}$s\\
  Gauss. Process  & 0.865 & 0.135 & 0.86 & 0.88 & 0.729 & $3*10^{-4}$s\\
  Decision Tree  & 0.879 & 0.121 & 0.88 & 0.88 & 0.757 & $4*10^{-7}$s\\
  Random Forest & 0.902 & 0.098 & 0.91 & 0.90 & 0.804 & $4*10^{-5}$s\\
  AdaBoosted DT & 0.849 & 0.151 & 0.84 & 0.87 & 0.698 & $7*10^{-6}$s\\
  XGboost & 0.820 & 0.180 & 0.79 & 0.88 & 0.643 & $2*10^{-5}$s\\
  LightGBM & 0.952 & 0.047 & 0.94 & 0.97 & 0.906 & $3*10^{-4}$s\\
  \verb|ikine.m| & N/A & N/A & N/A & N/A & N/A & $0.7583$ s\\
  \verb|ikine6s.m| & N/A & N/A & N/A & N/A & N/A & $0.0598$ s\\[1ex] 

\hline 
\end{tabular}
\label{tab:conf}
\end{table}

\begin{table}[htb]
\caption{Estimation net benchmark} 
\centering 
\begin{tabular}{c rrrrrrrr} 
    \hline\hline 
    Method  & MAE & MSE & EVS & R2 & Avg.Time\\ [0.4ex]
\hline 
  NN  & 0.0658 & 0.0095 & 0.8799 & 0.8787 & $2*10^{-4}$s \\
  Lin.Reg.  & 0.4106 & 0.2592 & 0.0794 & 0.0785 & $4*10^{-7}$s \\
  RidgeReg.  & 0.4106 & 0.2592 & 0.0796 & 0.0787 & $4*10^{-7}$s \\
  Decision Tree  & 0.5071 & 0.4556 & -0.5953 & -0.5962 & $6*10^{-7}$s \\
  Gauss. Process & 0.4905 & 0.3391 & 0.0 & -0.0012 & $8*10^{-4}$s \\
  k-NN  & 0.4303 & 0.2997 & -0.0617 & -0.0627 & $0.0531$s \\
   \verb|jacob0| (num.)  & N/A & N/A & N/A & N/A& $0.6146$ s\\
    \verb|jacob0| (ana.)  & N/A & N/A & N/A & N/A& $0.0626$ s\\[1ex] 

\hline 
\end{tabular}
\label{tab:est}
\end{table}

\begin{figure}
    \centering
    \includegraphics[width=.5\textwidth]{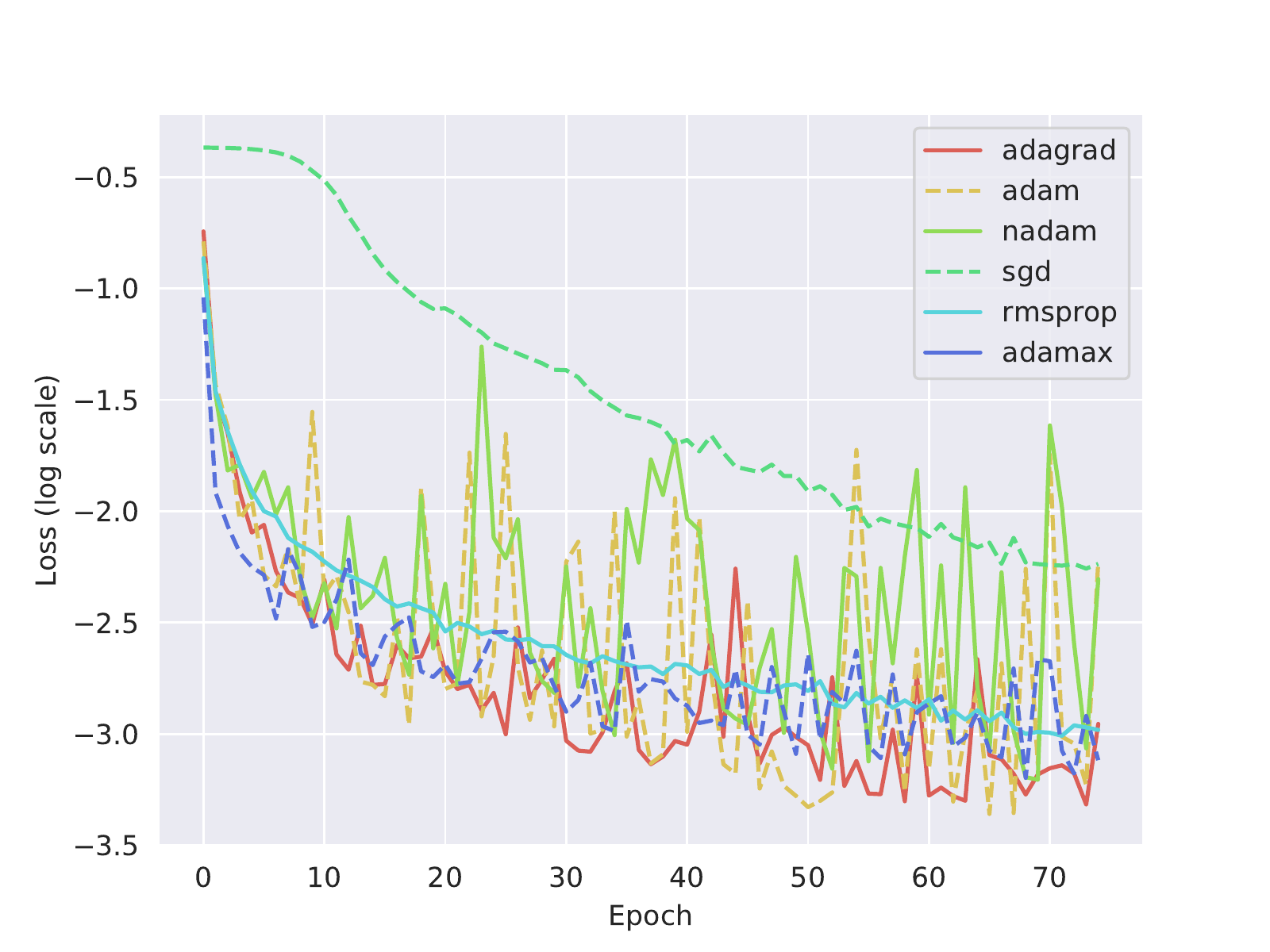}
    \caption{Comparison Between Training Gradient Curves Generated by Different Optimizers for Confidence Network}
    \label{fig:conf_compare}
\end{figure}

\begin{figure}
    \centering
    \includegraphics[width=.5\textwidth]{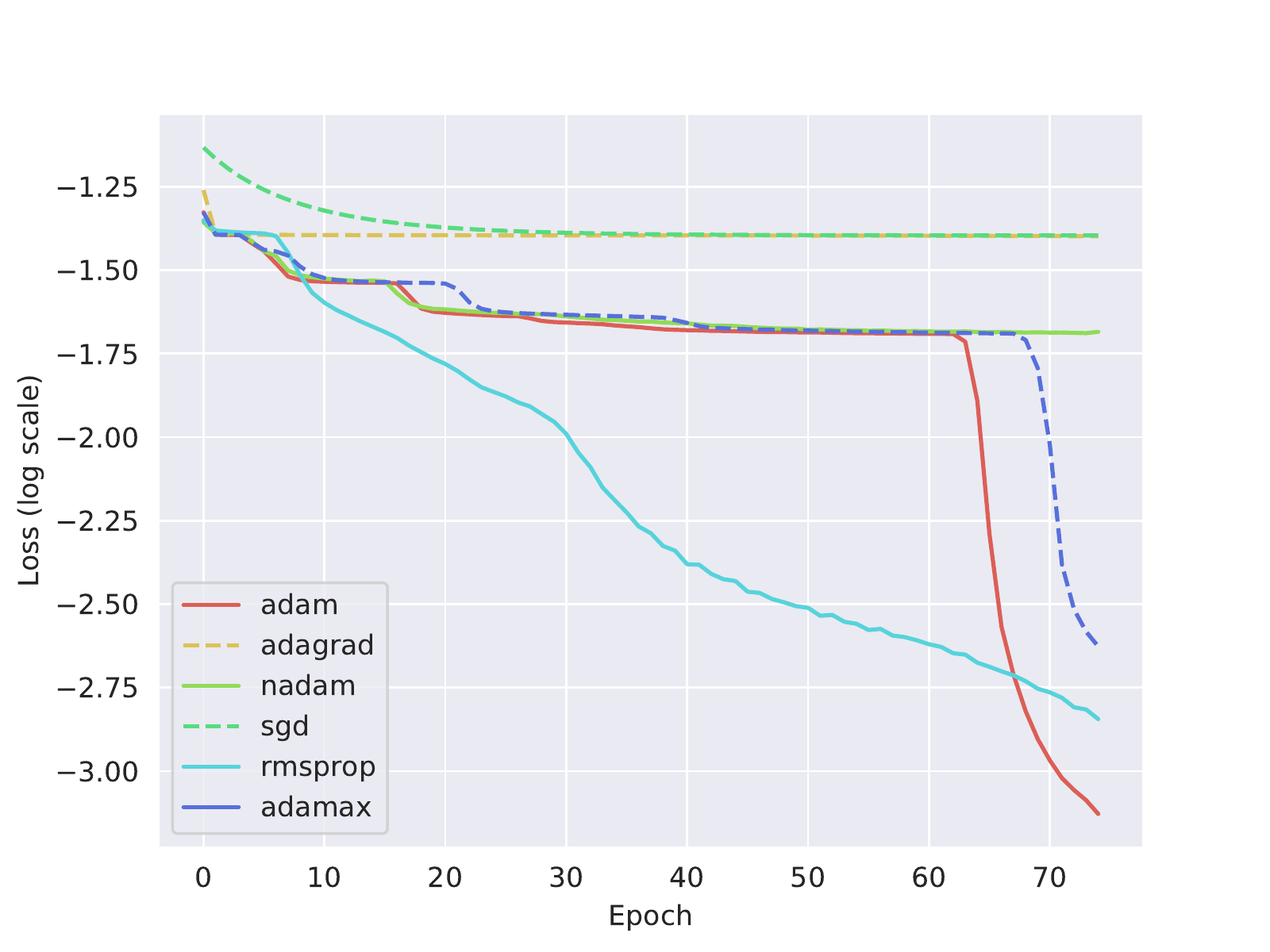}
    \caption{Comparison Between Training Gradient Curves Generated by Different Optimizers for Estimation Network}
    \label{fig:est_compare}
\end{figure}

We also compared network's training performance against different gradient step moethods. As demonstrated in Figure \ref{fig:conf_compare}, when training the confidence network, apparently Stochastic Gradient Descent\cite{GoodfellowDeep} ($\theta = \theta - \eta \cdot \nabla_\theta J( \theta; x^{(i:i+n)}; y^{(i:i+n)})$) has the most stable gradient curve without the apparent and dramatic rise and fall caused by the cycle model of training(freezing and unfreezing the parameters of the encoder). The best result and the worst result, given the training time of 100 epochs, are Adamax ($\theta_{t+1} = \theta_{t} - \dfrac{\eta}{\max(\beta_2 \cdot v_{t-1}, |g_t|)} \hat{m}_t$) optimizer and Stochastic Gradient Descent optimizer, both of which actually seem to be promising
leading to a final acceptable loss value. Other optimizers such as: 
\begin{itemize}
\item Adam\cite{Adam} $\theta_{t+1} = \theta_{t} - \dfrac{\eta}{\sqrt{\hat{v}_t} + \epsilon} \hat{m}_t$
\item Nadam\cite{Nadam} $\theta_{t+1} = \theta_{t} - \dfrac{\eta}{\sqrt{\hat{v}_t} + \epsilon} (\beta_1 \hat{m}_t + \dfrac{(1 - \beta_1) g_t}{1 - \beta^t_1})$
\item RMSprop\cite{RMSprop} $\theta_{t+1} = \theta_{t} - \dfrac{\eta}{\sqrt{E[g^2]_t + \epsilon}} g_{t}$
\end{itemize}
exhibit unstable gradient curves due to the training cycle and not converging really well.

The benchmark between different optimizers for the training of the estimation network is drastically different from that of confidence network, according to Figure \ref{fig:est_compare}. For the training of the estimation network, RMSprop\cite{RMSprop}, Adamax, and Adam\cite{Adam} all exhibit enough gradient to finally converge at an acceptable loss value with RMSprop being the most stable. The optimizers that performed well in the training of the confidence network including Nadan\cite{Nadam}, Adagrad\cite{Adagrad} and Stochastic Gradient Descent\cite{stochasticgradientdescent}, performed poorly with estimation network and did not converge after 80 epochs of training.
\subsection{Training Paradigms}
A specific way of training the network is employed to ensure the greatest accuracy when estimating the resulting Jacobian. The training of the combined model starts with pre-training the encoder network on the estimation network. Since the encoder network is essentially creating a regression, a $R^{2}$ value is calculated. However, because the decoder in estimation network is not tuned at this point of the training, the $R^{2}$ value would appear to be horrible.

Then in order to progress the training of the confidence network and estimation network at the same time without one being overfitting the encoded representation, we propose the \verb|cycle| model for the training. A \verb|cycle| is defined as training the confidence network two times and the estimation network one time since it takes a longer time for the confidence network to converge compared to the estimation network.

To prevent the training of the confidence network to affect the parameters of the already trained encoder network such that the encoder network does not bias toward the result of the confidence network, we freeze the encoder during the second step of the training where we fit the confidence network to the pre-trained encoder network. After the training, $R^{2}$ value for the confidence network is evaluated, and the network is further fine-tuned and compiled.

\begin{figure}[h]
    \centering
    \includegraphics[width=.5\textwidth]{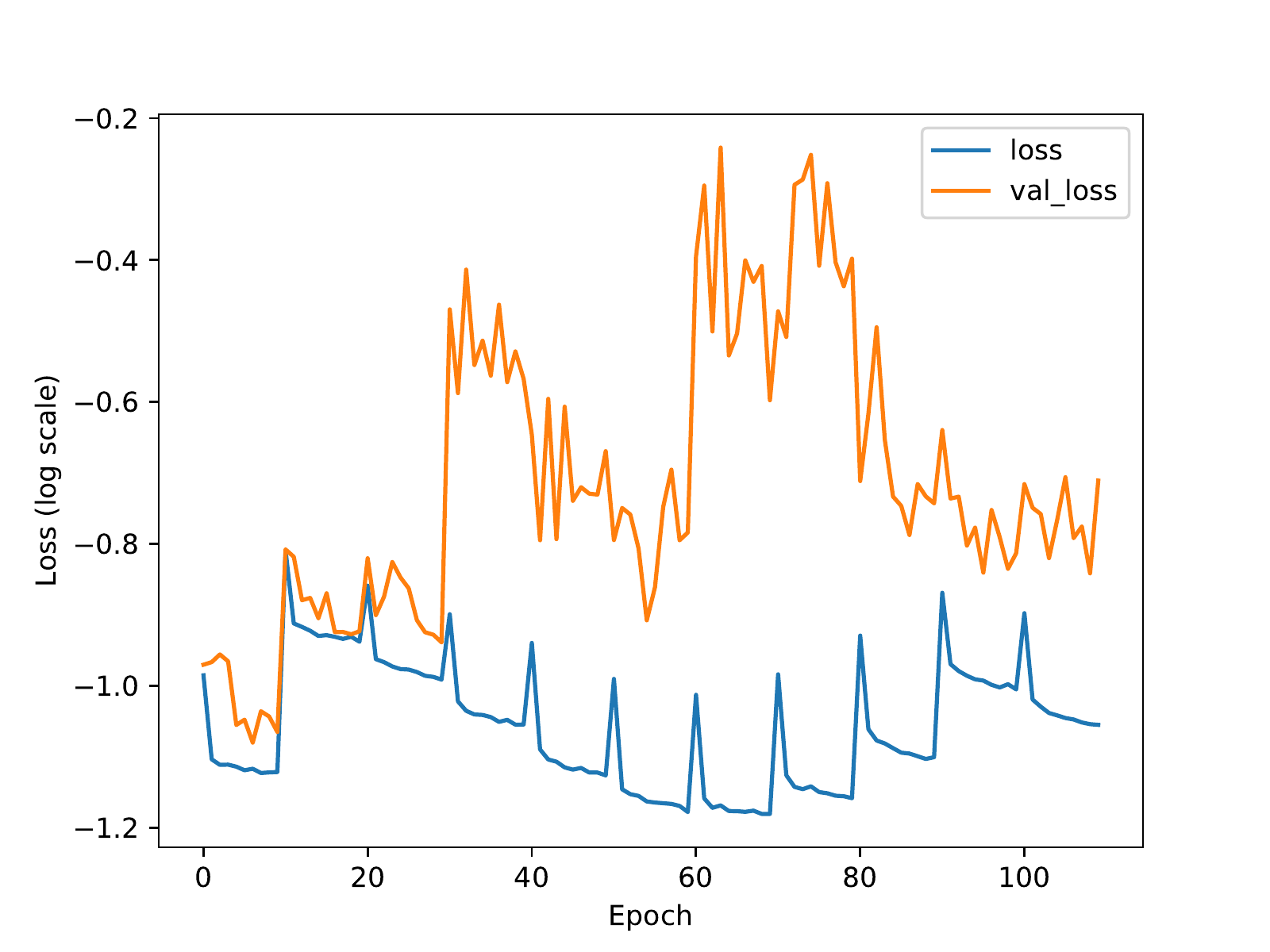}
    \caption{Confidence Network Gradient Curve}
    \label{fig:conf_history}
\end{figure}

\begin{figure}
    \centering
    \includegraphics[width=.5\textwidth]{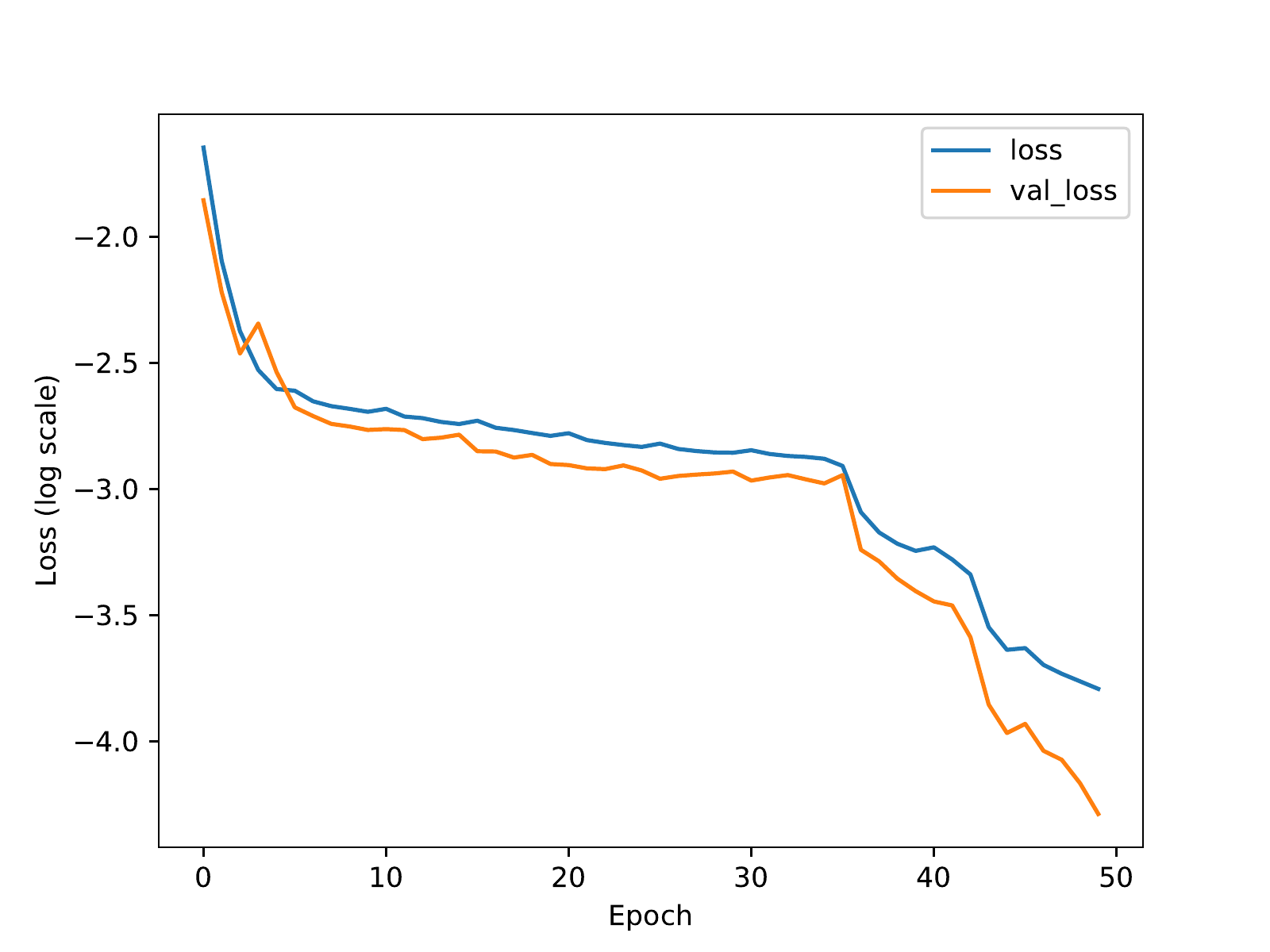}
    \caption{Estimation Network Gradient Curve}
    \label{fig:est_history}
\end{figure}

Then as we are training the estimation network, the encoder network is, and its parameters are adjusted along with those of the estimation network.
At the end, the combined model's $R^{2}$ value is evaluated. The cycled fashion of training the combined model contribute to the gradient curves observed in Figure \ref{fig:conf_history} and Figure \ref{fig:est_history}.

\subsection{Simulation on PUMA560 Robotic Manipulator}
To demonstrate the effectiveness of the neural network, we used visulization APIs from the Robotics Toolbox \cite{CorkeRobotics} to compare the results between the ground truth (generated by M3 library functions) and the predicted outputs. From Figure \ref{fig:p560} we can see that the results only vary slightly.

\begin{figure}
    \centering
    \includegraphics[width=.5\textwidth]{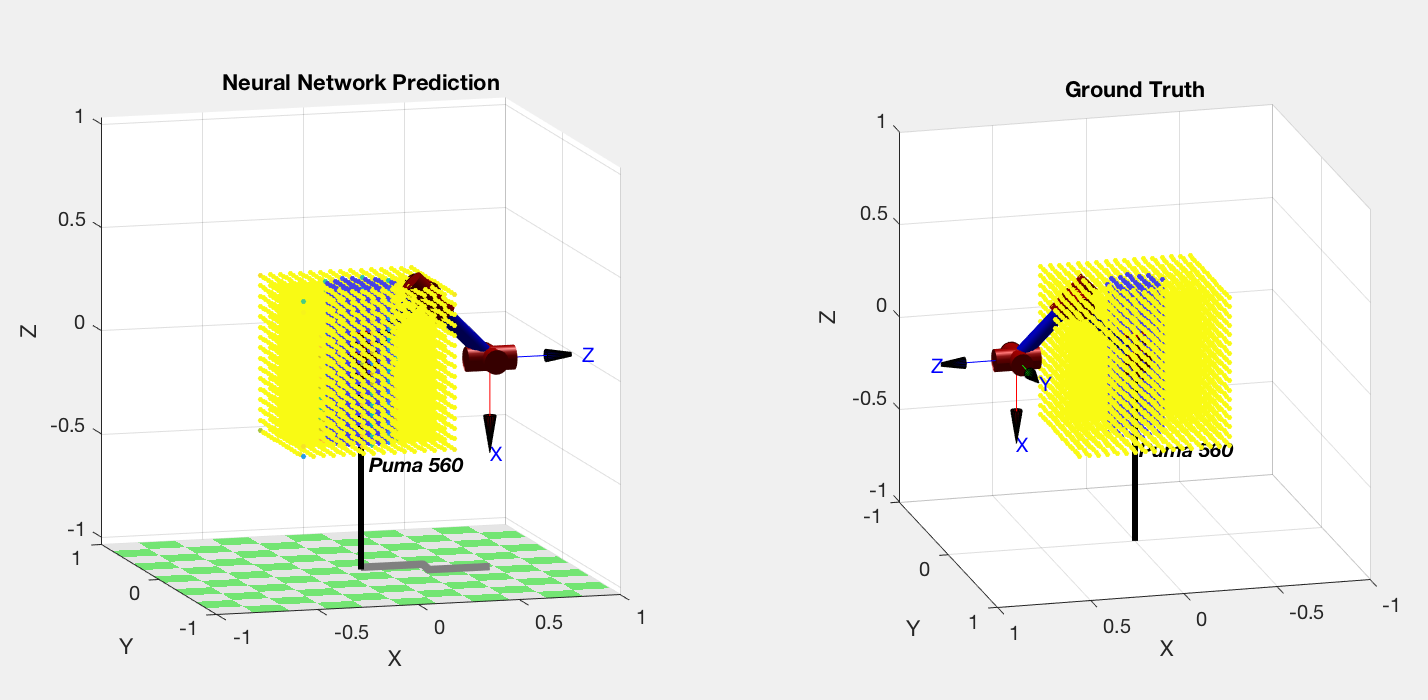}
    \label{fig:p560}
    \caption{Neural network prediction versus ground truth}
\end{figure}
\section{Future Works}
The paradigm of confidence-estimation network collaboration can be applied to different variants of the Jacobian estimation problem, like estimating its elements based on joint angle configurations rather than poses, which can then be used to create resolved rate motion control. Though this proposed framework is also suitable for the task, it still needs a forward kinematics function to obtain the end-effector pose from the joint angles, and further investigation is needed before a production-ready model can be created.
\begin{figure}
    \centering
    \includegraphics[width=.5\textwidth]{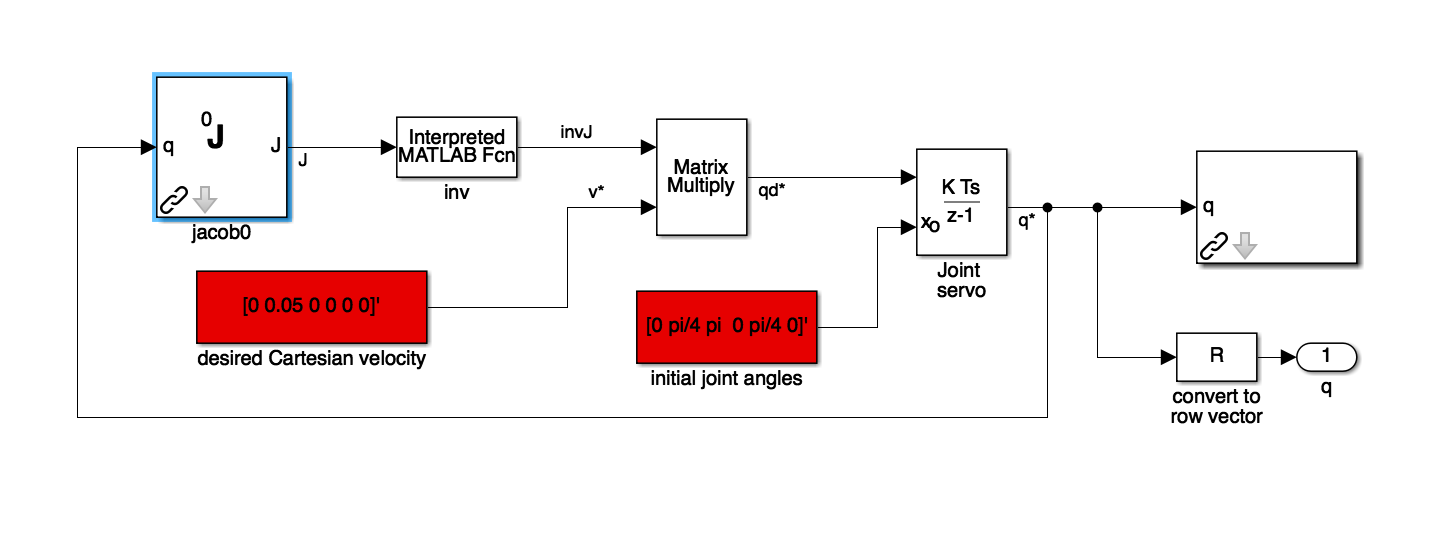}
    \label{fig:p560}
    \caption{Current resolved rate motion control scheme, provided by \cite{CorkeRobotics}. Note that the mapping is from joint angles $\textbf{q}$ to the Jacobian matrix in world frame $\textbf{J}$.}
\end{figure}

\begin{figure}
    \centering
    \includegraphics[width=.5\textwidth]{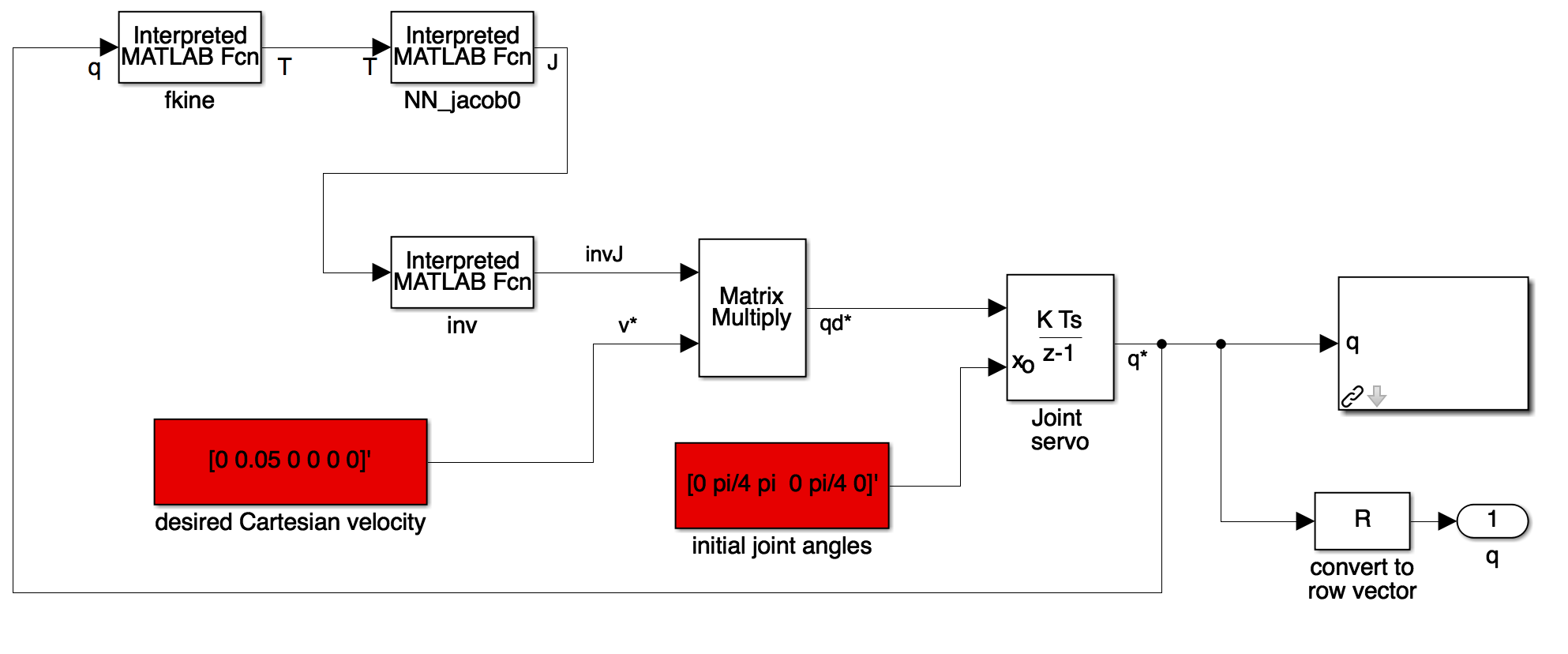}
    \label{fig:p560}
    \caption{Application of estimation net (trained on world frame dataset) in resolved rate motion control. The joint angles are first converted to end-effector pose $\textbf{T}$; then it is fed into the neural network for output.}
\end{figure}
\section{Conclusion}
In this work, we propose an accurate deep-learning framework that can generate the full workspace of serial-link manipulators by estimating its Jacobian matrix (given pose) and computing the confidence of the estimation. The architecture consists of an estimation network that approximates the Jacobian, either in world frame or in the end-effector frame and a confidence network that measures the confidence of the approximation. M3 (Manipulability Maps of Manipulators) is also introduced; it is a MATLAB robotics library based on Peter Corke's Robotics Toolbox, and it is used to generate the datasets for the neural networks constructed. Results have shown that not only is the proposed network is superior concerning runtime and portability when compared to numerical inverse kinematics, it is also more accurate than other machine learning alternatives.

\bibliographystyle{IEEEtran}
\bibliography{./references.bib}

\end{document}